%% file: sample-sigconf.tex
\documentclass[sigconf]{acmart}

\AtBeginDocument{%
  \providecommand\BibTeX{{%
    \normalfont B\kern-0.5em{\scshape i\kern-0.25em b}\kern-0.8em\TeX}}}

\newcommand{\sssec}[1]{ {{\flushleft \textbf{#1}}}}
\newcommand{\name}{Metamoran}

\usepackage{svg}
\usepackage{enumitem}
\usepackage[ruled,vlined,linesnumbered]{algorithm2e}

\acmConference[Woodstock '18]{Woodstock '18: ACM Symposium on Neural
  Gaze Detection}{June 03--05, 2018}{Woodstock, NY}
\acmBooktitle{Woodstock '18: ACM Symposium on Neural Gaze Detection,
  June 03--05, 2018, Woodstock, NY}
\acmPrice{15.00}
\acmISBN{978-1-4503-XXXX-X/18/06}

\begin{document}

\title{A Hybrid mmWave and Camera System for\\ Long-Range Depth Imaging}

\settopmatter{printacmref=false} 
\renewcommand\footnotetextcopyrightpermission[1]{} 
\pagestyle{plain} 

\author{Akarsh Prabhakara}
\affiliation{
\institution{Carnegie Mellon University}
\country{}}
\email{aprabhak@andrew.cmu.edu}
\authornote{Co-primary authors}

\author{Diana Zhang}
\affiliation{
\institution{Carnegie Mellon University}
\country{}}
\email{dianaz1@andrew.cmu.edu}
\authornotemark[1]

\author{Chao Li}
\affiliation{
\institution{Carnegie Mellon University}
\country{}}
\email{chaoli2@andrew.cmu.edu}

\author{Sirajum Munir}
\affiliation{
\institution{Bosch Research and Technology Center}
\country{}}
\email{sirajum.munir@us.bosch.com}

\author{Aswin Sankaranarayanan}
\affiliation{
\institution{Carnegie Mellon University}
\country{}}
\email{saswin@andrew.cmu.edu}

\author{Anthony Rowe}
\affiliation{
\institution{Carnegie Mellon University}
\country{}}
\email{agr@ece.cmu.edu}

\author{Swarun Kumar}
\affiliation{
\institution{Carnegie Mellon University}
\country{}}
\email{swarun@cmu.edu}


\begin{abstract}
mmWave radars offer excellent depth resolution even at very long ranges owing to their high bandwidth. 
But their angular resolution is at least an order-of-magnitude worse than camera and lidar systems. Hence, mmWave radar is not a capable 3-D imaging solution in isolation. We propose \name, a system that combines the complimentary strengths of radar and camera to obtain accurate, high resolution depth images over long ranges even in high clutter environments, all from a single fixed vantage point. \name\ enables rich long-range depth imaging with applications in security and surveillance, roadside safety infrastructure and wide-area mapping. Our approach leverages the high angular resolution from cameras using computer vision techniques, including image segmentation and monocular depth estimation, to obtain object shape. Our core contribution is a method to convert this object shape into an RF I/Q equivalent, which we use in a novel radar processing pipeline to help declutter the scene and capture extremely weak reflections from objects at long distances. We perform a detailed evaluation of \name's depth imaging capabilities in 400 diverse scenes.
Our evaluation shows that \name\ estimates the depth of static objects up to 90~m and moving objects up to 305~m and with a median error of 28~cm, an improvement of 13$\times$ compared to a naive radar+camera baseline and 23$\times$ compared to monocular depth estimation.


\end{abstract}

\maketitle

\input{intro-4} 

\input{related}

\input{method}
\input{depth_estimation}

\input{depth_imaging}
\input{partial_occulsions}

\input{results}

\input{limitations}

\bibliographystyle{ACM-Reference-Format}
\bibliography{citations_short}

\end{document}

%% file: intro-4.tex
\section{Introduction}
Modern surveillance systems are tasked with detecting objects of interest and observing if they violate any regulations such as perimeter restrictions. Moving beyond short range applications where depth cameras thrive \cite{zed2}, we ask the question, "what does it take to build a single vantage point sensing solution that can create depth images of objects at  ranges exceeding  100 meters?" A single vantage point solution allows for quick deployment in scenarios where infrastructure is hard to come by, with minimal calibration. For example, one can imagine a single pole-mounted platform that monitors people or cars trespassing large private areas or drones entering no-fly zones. State-of-the-art monocular camera based solutions (e.g: monocular depth estimation ~\cite{bhat2020adabins}) experience tens of meters of error for objects beyond $\sim$30~m. Lidar, although marketed to operate up to 100~m \cite{velo}, fails to detect certain objects between 30-50~m depending on object reflectivity characteristics and ambient sunlight.

Apart from cameras and lidars, another mainstream depth sensor today is the mmWave radar. One of the most appealing features of mmWave radar systems arises from the high bandwidth of its operating spectrum, which enables object detection often as far as 150-300~m at cm-scale depth resolutions. This finds application in a wide range of areas, including security~\cite{bjorklund2012evaluation}, automobile safety~\cite{wenger1998automotive}, industrial sensing and control~\cite{dandu2021high}. Yet, mmWave radars 
suffer from poor angular resolution along both azimuth and elevation due to the limited number of antennas. The best commercially-available radars, at best, achieve an angular resolution of 1.5$^{\circ}$ \cite{tiradar}, which is at least 10$\times$ worse than cameras. This has led to mmWave radars being largely restricted to niche applications -- for instance, in airport security~\cite{bjorklund2012evaluation} or collision sensing~\cite{wenger1998automotive} ---  where their impressive operating range and depth resolutions are not fully utilized. 

We make a key observation that radars and cameras are mutually complementary to each other offering long range accurate depth estimates and high angular resolution semantic information respectively. This naturally leads us to the question: \textit{Can we fuse mmWave radar and camera to provide the best of both worlds and build a long-range high angular-resolution depth imaging solution?}
This paper presents \name\footnote{A fictional race from the Dragon Ball Universe that taught Son Goku the Fusion technique.}, a hybrid mmWave and camera-based sensing system that achieves high angular and depth resolution in high clutter environments for static and mobile objects up to 90m and 300m respectively. Efforts have been made to fuse radar and camera data in the past, primarily for imaging under physical \cite{palffy2019occlusion} or weather-related occlusions \cite{liu2021robust}, and for short range object detection and tracking \cite{chang2020spatial}. This paper instead considers the unique problem of hybrid mmWave radar and camera sensing for long-range depth imaging. This is difficult because unlike systems that operate over short ranges, the first peak detected in radar doesn't necessarily correspond to the detected object in the image.  This is primarily because of overwhelming reflections from ambient but out-of-interest objects that can clutter the scene.

\begin{figure*}
    \centering
    \vspace*{-0.1in}
    \includegraphics[width=0.9\textwidth, height=6.5cm]{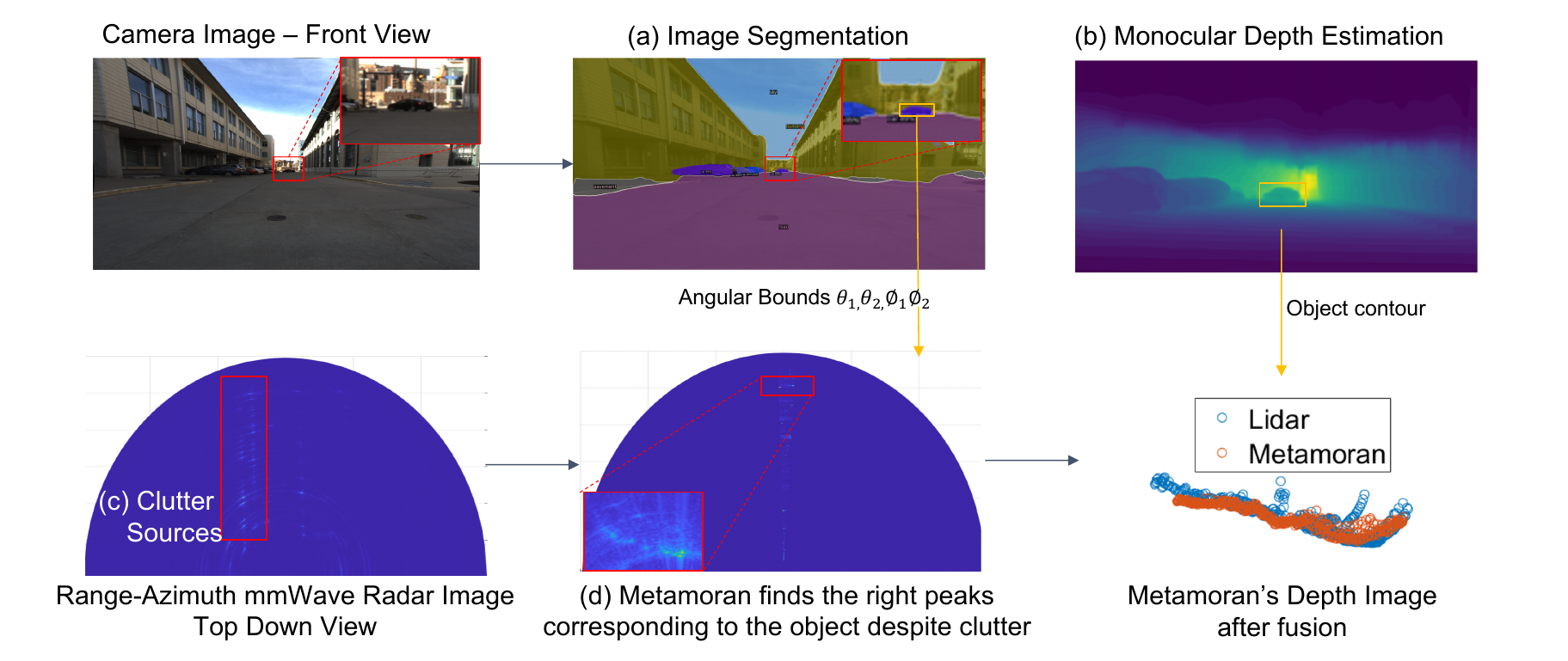}
        \vspace*{-0.1in}
    \captionof{figure}{\textbf{\name}\ produces high resolution depth-images up to 300~m away from objects-of-interest by co-optimizing radar processing with inputs from vision techniques such as image segmentation and monocular depth estimation. }
    \label{fig:main}
    \vspace*{-0.1in}
\end{figure*}

\name\ leverages  semantic information from the camera image to help the radar disambiguate between objects-of-interest while eliminating clutter. Our technical approach is a series of algorithms ranging from simple to complex to declutter the scene. Once the scene is decluttered and  objects-of-interest are identified in the radar image, \name\  uses the camera image again to synthesize a high angular-resolution depth image of the objects, including their internal depth variations. Finally, we present various solutions to tackle clutter arising due to partial occlusions or poor lighting, where camera performance suffers. 

\sssec{Accurate depth estimation in high clutter. } An intuitive starting point for \name\ to eliminate unnecessary clutter is to segment the camera image  (see Fig.\ \ref{fig:main}a) and use the radar to look for peaks \textit{only} within the angular span ($\theta_1, \theta_2, \phi_1, \phi_2$) occupied by the objects-of-interest as identified in the segmentation output. This helps the radar ignore reflections from out-of-interest objects. 

A practical challenge in designing this arises because of strong clutter from objects such as buildings, lamp posts and fences. Presence of a strong reflector creates side lobes that spreads across the angular axis (see Fig. \ref{fig:main}c). This implies that even after segmenting radar to only angles where the object-of-interest is present, side lobes from strong reflectors at other angles create ghost peaks within the angles-of-interest. This is extremely critical as depth estimation can have significant errors if ghost peaks are selected.

To tackle this challenge, we design a new radar processing algorithm which selects the correct peaks by leveraging the camera image. Our key idea is unlike other clutter, the peaks in radar images of an object-of-interest \textit{look like} the object. For example, the contour of a car is quite evident in Fig. \ref{fig:main}d. In other words, camera images of the scene tell us the precise shape of the contour of a car we can expect in radar images.  \name\ therefore designs an I/Q \textit{contour} template that mimics the precise signal structure we would find in the radar image, synthetically, based solely on information in the camera image. The availability of the camera image is crucial here to let us do this -- it tells us the object type and angles-of-interest, and objects' internal depth variation, although not on an absolute scale. We further describe various optimizations that use the so far discarded, out-of-interest image segment to identify undesired strong reflectors and create a \textit{clutter} template to eliminate the side lobes. Sec.~\ref{sec:depthestimation} describes the specifics of our  \textit{contour} and \textit{clutter} template design and how they help weed out spurious peaks to obtain accurate depth estimates for an object.

\sssec{Depth imaging. } At this point, we know which peak or a set of sparse peaks correspond to the object and, thereby, have a good estimate of the absolute depth of the object. However, because of poor radar angular resolution, we do not have a high angular resolution depth image which shows how depth varies across angles spanned by the object. To account for this, we leverage monocular depth estimation  (see Fig. \ref{fig:main}b) which provides depth estimates at the high angular resolution of a camera. Although the \textit{absolute} depth estimates of these algorithms have several meters of error, we make an important observation that the \textit{relative} depth variation within an object-of-interest (such as a car) is captured to a sufficient degree of accuracy by these algorithms. Thus, we fuse the sparse peaks from radar and dense point cloud from monocular depth estimates to get an accurate, high resolution depth image. Sec. \ref{sec:depthimaging} describes the challenges and details about implementing this fusion.

\sssec{Partial occlusions and vision impediments. } Next, we consider occlusions and various vision impediments. If an object-of-interest is completely occluded by significant blockage then neither the radar nor camera sees it. But, if they are partially occluded then radar perceives peaks both from object-of-interest and occluding object. This is again a problem of peak selection. We note that the image segmentation identifies both the object-of-interest and occluding object. Our system uses the segmentation mask to identify which of the two is at the foreground or the background. We then accordingly choose the peaks either closer or farther away.

Although mmWave radar is robust to different environmental conditions, a natural question to ask is what happens to our system when camera fails due to poor lighting or lack of visibility due to bad weather such as fog. While our decluttering strategies are only possible because of rich camera information, we design an alternative system that takes over during camera failures but only works in clutter-free scenes and at short ranges 0-20m. Section \ref{sec:occlusions} describes how this system that estimate depth, create a sparse image and classify the object type without using camera images.

We implement \name\ with a TI MMWCAS-RF-EVM radar with 1.5$^{\circ}$ azimuth resolution and a FLIR Blackfly S 24.5MP color camera. Due to the relative lack of public mmWave radar I/Q datasets over long distances, we collected extensive radar data ($\sim$400 static and dynamic scenes totalling 125 GB of I/Q samples) along with high resolution, raw camera images and lidar point clouds  in diverse scenes outdoors at a major U.S. city. \name's datasets and code will be \textbf{open sourced} to benefit the community. Our results show:
\begin{itemize}[leftmargin=*]
\item Depth estimation of static objects up to 90~m and mobile objects up to 305~m.
\item Accurate depth estimation with a median error of 28~cm for static objects-of-interest at distances of up to 60~m, in high clutter environments -- an improvement of 23$\times$ vs. monocular depth estimation and 13$\times$ versus a na\"ive camera + radar solution. 
\item System resilience to partial occlusions and camera failure.
\end{itemize}

\sssec{Contributions.} We make the following contributions: (1) A novel system that combines camera and mmWave sensing to achieve high resolution depth images. (2) A novel radar processing algorithm to eliminate clutter using image segmentation and monocular depth estimation. (3) A detailed implementation and evaluation of \name\ in various high-clutter environments to demonstrate substantial improvements in long range depth imaging.


%% file: related.tex
\section{Related Work}
{{\flushleft  \textbf{Wireless and Radar Depth Sensing: }}} Recent years have seen extensive work in sensing the environment through wireless imaging~\cite{Huang2014, DepatlaAPM17},  location tracking~\cite{SpotFi, mtrack} and material sensing~\cite{Wu2016,Tam}, with much of this work limited to ranges of few tens of meters. Some prior work has also explored high-resolution mmWave radar systems for through-wall/through-obstruction imaging~\cite{DepatlaAPM17, guan2020through}, security scanning~\cite{stanko2008active} and predictive maintenance~\cite{mitchell2020evaluation}. While complementary, these solutions are not designed to measure high-resolution depth images at extended distances, primarily due to the limited azimuth resolution of radar platforms. 

{{\flushleft  \textbf{Depth Sensing using Cameras/Lidar:}}} Cameras~\cite{mejias2010vision}, lidars~\cite{ramasamy2016lidar} and depth imaging~\cite{huang2017visual} are often used in diverse outdoor 3-D imaging applications. Some depth camera systems (e.g. monocular depth estimation~\cite{bhat2020adabins}) struggle at extended distances, some (e.g. stereo-vision~\cite{smolyanskiy2018importance}) require extended baselines for high accuracy,  while others (e.g. IR structured light~\cite{sarbolandi2015kinect}) function poorly under ambient light. More broadly, systems struggle to measure depth at a high resolution at long range, with about meter-scale accuracy at up to 80m range in monocular depth estimation cases \cite{Zhao_2020} and only operating up to around 20m in the case of depth cameras \cite{zed2}. Some lidar systems~\cite{poulton2019long} offer higher accuracy at extended ranges, however face other significant limitations stemming from the power consumption of the laser as well as robustness to dust, weather conditions and coexistence with other lidar platforms~\cite{bijelic2018benchmark,kim2015occurrence}.


{{\flushleft \textbf{RF-Camera Fusion:}}} Camera and RF fusion has been proposed for automatic re-calibration \cite{xu2012towards}, industrial workplace \cite{savazzi2015device}, localization~\cite{alahi2015rgb}, person identification~\cite{fang2020eyefi} and fall detection \cite{kianoush2016device}.  Radar-Camera fusion has also been studied for diverse vehicular applications including attention selection to identify objects-of-interest~\cite{Ji2008,chang2020spatial}, tracking mobile objects~\cite{long2018fusion,sengupta2019dnn,zhang2019extending} better object perception and classification under poor weather~\cite{Kato2002, john2019so, han2016frontal}, detecting vehicles and guard rails~\cite{steux2002fade, alessandretti2007vehicle, Ji2008} and generating obstruction-resilient 2D images~\cite{lekic2019automotive}. Vision-based sensing has also been used for more effective communication using mmWave \cite{searchlight, nishio}. Beyond radar and vision,  prior work has used multi-modal fusion across a variety of sensors for tracking human activity~\cite{Li2017}, autonomous driving~\cite{Cho2014} and beyond. We distinguish ourselves from this work by combining mmWave radars and camera for high-resolution depth imaging at long ranges, including under partial occlusions.

%% file: method.tex
\section{mmWave Radar Primer}\label{sec:primer}
Radars, once only limited to military applications, are today used ubiquitously in a variety of applications from  airport security~\cite{bjorklund2012evaluation}, automotive applications~\cite{tokoro1996automotive}, human-computer interfaces~\cite{liu2020real} and industrial automation~\cite{mazgula2020ultra}. A key factor which enabled this trend was the usage of mmWave frequencies which allowed for compact antenna arrays and wide bandwidths, both of which are crucial for radars' target ranging and imaging capabilities. mmWave radars, as the name suggests, use radio waves of millimeter scale wavelengths in either 60~GHz or 77-81~GHz by first actively illuminating an environment and then processing the reflections from various objects in the environment. This is noticeably different from modern image sensors which purely rely on passively sensing rays which make their way to the sensor. The reflections from the objects encode useful information such as objects' range, azimuth, elevation and velocity with respect to radar. The transmitted illumination and radar hardware are the main factors which limit the radars' ability to generate high resolution depth images of the scene.

\sssec{Advantages of mmWave Radar: } Most commodity radars transmit a Frequency Modulated Continuous Wave (FMCW) signal which continuously changes its frequency over time to span a significant bandwidth $B$. A radar's range resolution is fundamentally limited by this effective bandwidth of the transmitted signal as $\frac{c}{2B}$ ($c$ is speed of light). In the 77~GHz band, we have a theoretical range resolution of 3.75~cm over tens of meters. In this regard, radars are on par with time of flight lidars which report a similar range accuracies. However, unlike lidars, radars work in all weather conditions (rain, snow, fog) and extreme ambient lighting (sunlight)~\cite{mullen2000hybrid}. Commercial time of flight lidars advertise a maximum range of 100~m but depending on ambient sunlight and object reflectivity characteristics they can stop detecting objects even at 30-50~m. In contrast, radars today have low enough noise floor to detect objects such as cars and person between 150-300~m too. Moreover, while most commercial lidars are still time of flight based, FMCW radars can do Doppler processing to detect moving objects much weaker than surrounding clutter much more easily. These attributes favor mmWave radars for long range depth estimation.

\sssec{Limitations of mmWave Radar: } However, radars unfortunately have worse azimuth and elevation resolutions compared to both cameras and lidars. While range resolution is limited by the bandwidth of the radar signal, angular resolutions are dictated by the number of antenna elements that are packed on a radar. As the number of antenna elements increases, the resolution improves. The best state-of-the-art commercial mmWave radar available~\cite{tiradar} with as many as 86x4 antenna elements has a 1.4$^{\circ}$x18$^{\circ}$ angular resolution. 77~GHz radars with high elevation resolution aren't available today as they are mostly used for automotive use cases where elevation is not very important. The focus of this paper is in enhancing the azimuth resolution and generating 2D depth images (depth vs azimuth). In contrast to radar, state of the art lidars today achieve 0.1$^{\circ}$x2$^{\circ}$, at least 10$\times$ better angular resolution than radars~\cite{liu2008airborne}. With a poor angular resolution, radar images look very coarse and blobby in the angular domain. While more antenna elements can be added, they come at significant increases in device cost and form-factor -- bridging the 10$\times$ gap is simply not an option with today's state-of-the-art hardware. Even commodity cameras, because of their dense focal planar array image sensors, are better than radars in terms of angular resolution at about 0.02$^{\circ}$x0.02$^{\circ}$~\cite{napier2013cross}. This observation leads us to study combining the high angular resolution of camera systems with the high depth resolution of mmWave radar at long ranges -- an approach we describe in subsequent sections.

\section{\name's Approach}
\name\, at a high level, takes as input camera and 77~GHz mmWave radar data from a scene. We use these inputs to fuse and return a high-resolution depth image for specific objects-of-interest at distances of several tens of meters. We specifically consider cars and people as objects-of-interest -- key to surveillance applications. We also consider traffic signs as a sample from a class of objects that are always static. 

\subsection{Image Pre-Processing}\label{sec:semseg} 

\name's first step is to pre-process camera image data to learn about the approximate span in azimuth and elevation of objects-of-interest and obtain a high angular resolution depth image using camera alone. As mentioned in Sec.~\ref{sec:primer}, we exploit the high angular resolution of camera systems that are at about 0.02$^{\circ}$x0.02$^{\circ}$~\cite{napier2013cross} --  orders-of-magnitude better than mmWave radar systems. \name's vision pre-processing steps below are therefore crucial in providing prior information on the angular location of objects-of-interest and the depth variation within an object to first help declutter the scene and then help synthesize an accurate, high resolution depth image.


\begin{figure}
\centering
\includegraphics[width=0.8\columnwidth]{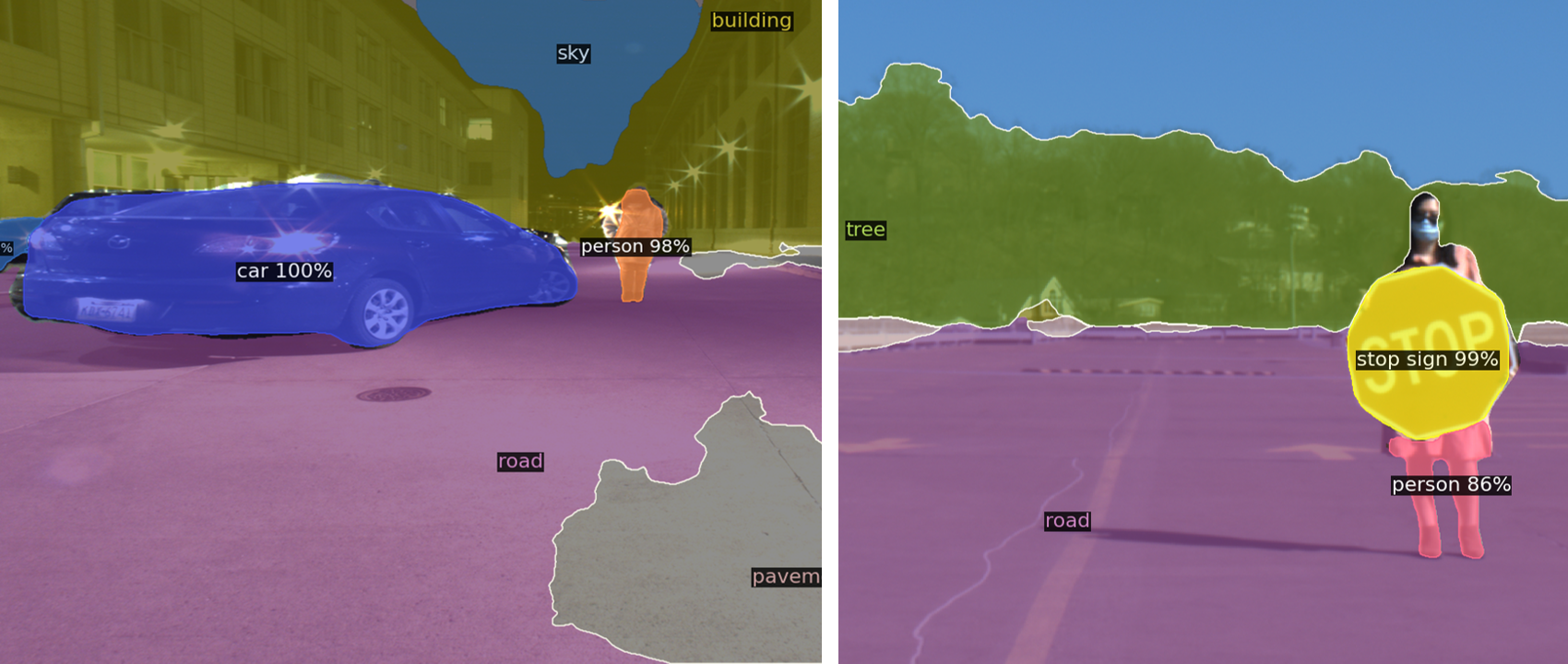}
\vspace*{-0.1in}
\caption{\textbf{Image Segmentation:} \name\ uses image segmentation to identify the angular bounds along the azimuth and elevation axes for objects-of-interest -- cars, persons, traffic signs --  with semantic labels assigned. }
\vspace*{-0.2in}
\label{fig:panoptic}
\end{figure}

\sssec{Image Segmentation: } To find the angular bounds of objects-of-interest, we perform state-of-the-art image segmentation which labels objects by their type and creates masks that capture the outline of these objects (see Fig.~\ref{fig:panoptic} for an example). We perform image segmentation using Detectron2 \cite{wu2019detectron2} trained with KITTI dataset. This model has been previously trained on several objects including cars, persons and traffic signs in various environments. We use these types of objects as our primary test subjects without additional model tuning. This image segmentation combines the best of both worlds from semantic segmentation and instance segmentation, by providing a segmentation mask (outline), a semantic label for the mask and instance ID for each detected semantic object as shown in Fig.~\ref{fig:panoptic}. This is vital as many of our scenes are in high clutter environments, which for example, in addition to a car of interest can have several other cars. The instance ID helps us isolate the car that we are interested in. The segmentation mask directly provides the angular bounds that is fed into radar processing pipeline to declutter the scene in Sec.~\ref{sec:depthestimation} below.

\sssec{Monocular Depth Estimation: } Next, we perform state-of-the-art monocular depth estimation. We use this scheme both as a baseline for comparison and to provide an approximate depth variation across the angles spanned by the object. We use AdaBins~\cite{bhat2020adabins} for monocular depth estimation and note that state-of-the-art monocular depth estimation is poor in terms of accuracy at extended distances, with errors of about 19.5~m for objects that are 60~m away (see Fig. \ref{fig:depthrange}). Nevertheless, we see that monocular depth estimation provides useful prior information on the approximate depth variation within an object. Combined with image segmentation it provides a rough 3-D shape (outline) of the object that serve as inputs for our decluttering algorithm Sec.~\ref{sec:depthestimation} and high resolution imaging algorithm in Sec. ~\ref{sec:depthimaging}.

\begin{figure}
\centering
\includegraphics[width=\columnwidth]{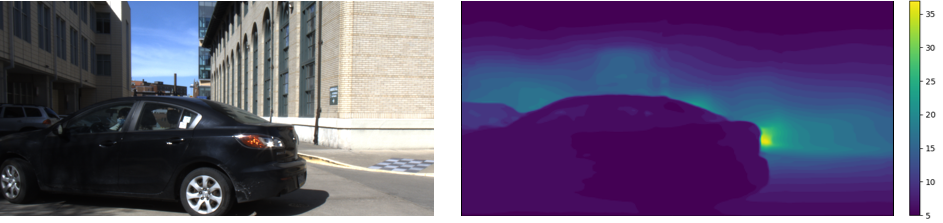}
\vspace*{-0.2in}
\caption{Monocular depth estimation gives a high angular resolution depth image which is promising for fusing with \name's sparse point clouds.}
\vspace*{-0.25in}
\label{fig:monoculardepth}
\end{figure}





\subsection{System Architecture and Outline}

Fig.~\ref{fig:main} depicts the architecture of our system that we elaborate upon in the following sections. After applying the image pre-processing steps of image segmentation and monocular depth estimation, we perform the following. (1) \textbf{Accurate depth estimation in high clutter:} We design a novel radar processing algorithm in Sec.~\ref{sec:depthestimation}
that uses priors from image segmentation and monocular depth estimation to declutter the scene and obtain an accurate depth estimate for each object-of-interest. (2) \textbf{Depth imaging:} We then show how to synthesize a depth image by fusing accurate depth estimate obtained so far and high angular resolution monocular depth estimates (Sec.~\ref{sec:depthimaging}). (3) \textbf{Partial occlusions and vision impediments:} Our final step (Sec.~\ref{sec:occlusions}) is to build resilience to partial occlusions and vision-based impediments.

%% file: depth_estimation.tex
\section{Accurate Depth Estimation} \label{sec:depthestimation}

\begin{figure}
\centering
\includegraphics[width=\columnwidth]{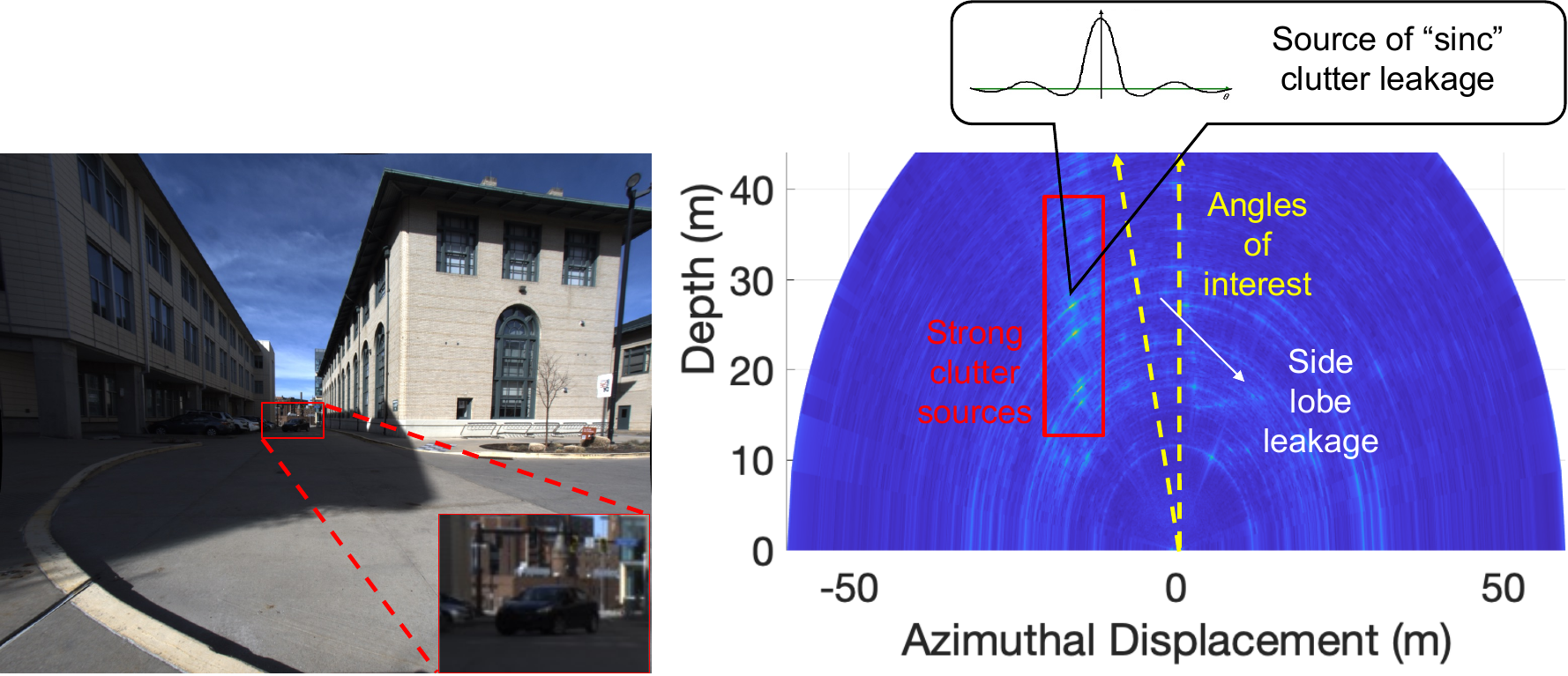}
\vspace*{-0.1in}
\caption{\name\ tackles the overwhelming clutter introduced by strong reflectors such as buildings which dwarf the reflection from object-of-interest.}
\vspace*{-0.2in}
\label{fig:clutter_source}
\end{figure}

To find the object-of-interest, \name\ first uses the image segmentation mask and segments the radar image. For example, if a car lies between $-5^\circ$ and $0^\circ$, the radar heatmap seen in Fig \ref{fig:clutter_source}, is truncated to these angular limits. Assuming that the object is in line of sight with respect to radar, in an ideal world, within the angular limits there should only exist peaks corresponding to the object-of-interest and nothing else. However, in high clutter environments, strong, out-of-interest reflectors which can even lie outside the angular limits, tend to leak their signal into angles-of-interest (see Fig. \ref{fig:clutter_source}). Such strong reflectors like buildings, tend to spread out their signal along the azimuth axis in a sinc-like fashion with decreasing side lobe levels. These side lobes affect across all angles at the same range bin and show up as false peaks within the angle-of-interest. A naive radar camera fusion would end up choosing these false peaks. The rest of this section describes our approach in accurately detecting peaks even in the presence of high clutter.


\subsection{Computing Object Depth}\label{sec:depth}

\begin{figure}
\centering
\includegraphics[width=\columnwidth]{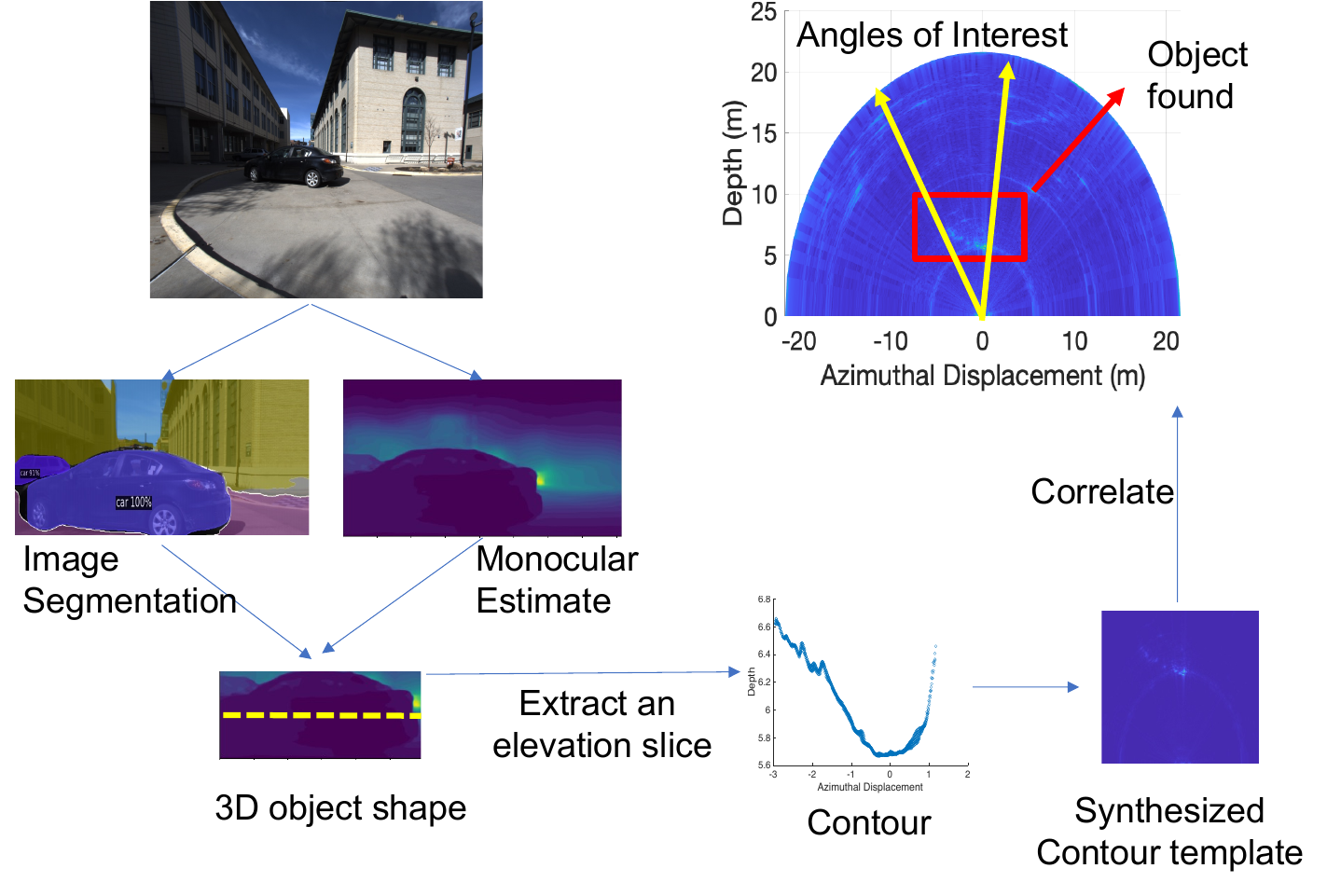}
\vspace*{-0.25in}
\caption{\name\ uses information from camera image segmentation and monocular depth estimation to obtain a coarse contour of the object-of-interest. It then uses this contour to perform correlation to find the object in radar image amidst clutter and thereby estimate its depth accurately.}
\vspace*{-0.2in}
\label{fig:clutterleakage}
\end{figure}

After segmenting the radar image to the desired angles-of-interest, \name's key next step is a novel radar processing algorithm, which searches for peaks that resemble the shape of the object. Our idea is to build an approximate shape of the object by leveraging monocular depth estimates  that capture the relative depth variation of all objects in the entire scene in an RGB-D depth image as explained in Sec. \ref{sec:semseg} (see Fig.~\ref{fig:clutterleakage} for an example). We then extract a portion of the monocular depth estimate RGB-D of only the object-of-interest based on its spatial extents obtained from semantic segmentation. Intuitively, this portion captures the 3-D shape of the object-of-interest. Because our radar has no information about elevation, we take an elevation slice from the segmented depth image. We call this a \textit{contour}. We note two important properties of a contour. First, it invariably captures the internal variations of depth within the object much better than radar -- mainly because cameras have denser pixels (i.e. resolution along the azimuth and elevation  axes) versus radar. Second, the contour is likely significantly erroneous in terms of where it believes the entire objects' absolute depth is, compared to the radar. As stated earlier, this is because a radar's absolute depth estimates are far superior to camera-based techniques, particularly for far-away objects.  

\begin{algorithm}[!b]
    \SetKwInOut{Input}{Input}
    \SetKwInOut{Output}{Output}
    \caption{Depth Estimation Algorithm}
    \Input{Image Segmentation Object Mask, $P$
    \\Monocular Depth Estimation, $M$ 
    \\Raw I/Q Radar capture, $h$}
    $S$ = $M \cdot P$ \hfill\tcp{ Approximate 3D shape of object}
    $C(x,z)$ =  \textsc{GetShapeContour}$(S(x,y,z))$\\
    \For{depth $d$}{$h_{template}^{d} = $\textsc{ShiftByDepth}$(C(x, z), d)$\\
 $P(d) = corr(h_{template}^{d},h)$ \hfill\tcp{Matched Filtering}}
    $d^{*} = \underset{d}{\operatorname{argmax}}  P(d)$\hfill\tcp{Depth Estimate}
    \Output{$d^{*}$\hspace*{0.3in}\hfill \tcp{Depth Estimate}}
    \label{algm:alg1}
\end{algorithm}

At this point, \name\ attempts to explore at what depth signals from the contour exist within the received radar images. To estimate this quantity, our radar processing algorithm, at a high level, uses the extracted contour to synthesize an I/Q \textit{contour template} -- i.e. a simulated synthetically generated radar reception of an object that has the exact geometry as defined by the contour. If the contour is moved to different absolute depths and compared against the received signal, it will ultimately coincide with the true peaks corresponding to the object-of-interest.  More formally, by simply correlating, shifted versions of the template with the original signal, we can find the true depth accurately. Algorithm 1 and Fig. \ref{fig:clutterleakage} captures the key steps in our depth estimation algorithm. 



\sssec{Detailed Algorithm: } Let $P$ be the binary mask which corresponds to the location of the object-of-interest in the image. The monocular depth estimate obtained can be captured in a matrix $M$, which is basically the "D" slice from RGB-D image. We can then obtain the approximate 3D shape $S$ of the object by element wise multiplication of $M$ and $P$. $S$ is now a matrix which is largely 0, but in pixels where object is present, it has monocular depth estimates. Rather than using the full 3D shape, for our 2D radar, we extract a contour $C$ as essentially a row chosen from $S$. The row index translates to elevation angle and if radar and camera are co-located, we simply choose the centermost row. The column index translates to azimuth angle. We convert the column indices to appropriate azimuth angles. Choosing non-zero elements in this row, we have a point cloud that can be indexed by azimuth angle and depth value. We can then transform these coordinates to $C(x,z)$.

Using $C$, \name\ synthesizes the contour template by modeling each point on the contour $C(x,z)$ as a point reflector. In its simplest form, one can obtain this point's contribution to the synthesized FMCW signal as \cite{tifunda}:
\begin{equation}
h_{template; i}(n) = \alpha e^{j \frac{4 \pi D_{i}}{\lambda}} e^{j 2 \pi \frac{D_{i}}{D_{max}} n}
\label{eqn:1}
\end{equation}

where, $\alpha$ is the amplitude of the received signal, $D_i$ is the distance between $(x_{i},z_{i})$ and the radar antenna, $D_{max}$ is the maximum distance that the radar is configured to operate, and $n$ indexes ADC samples.

Superposing each point's contribution we obtain the overall signal template for the entire contour as $h_{template}$. Each time we shift the point cloud to $d$, we synthesize a new template $h_{template}^{d}$. \name\ then applies a matched-filter to obtain $P(d)$ -- the correlation of the contour template at each possible depth $d$ relative to the radar by processing the received signals across frequencies. Mathematically, if $h$ is the original received radar signal, we have:
$$P(d) = corr(h_{template}^{d}, h)$$
We then report the depth estimate of this object as the value of $d$ that corresponds to the maximum of $P(d)$, i.e.
$$d^* = \arg \max_d P(d)$$ 

With $d^{*}$, we know accurately the closest depth of the object with respect to the radar. The effective depth imaging of an object relies first on accurate peak estimation corresponding to the object. In this subsection, we showed that by using information from segmentation and monocular depth estimate, we can pick the objects' peaks accurately. To further help finding object peaks in cluttered conditions, the following subsection describes how camera information can also be used to suppress the clutter.

\subsection{Clutter Suppression}
\label{sec:clutter}

Clutter due to strong reflections from undesired objects can impede \name. For instance, even if an undesired object is at an azimuth significantly different from the desired object, it's side lobes can create ghost peaks that causes interference. Worse still, some reflectors may be orders of magnitude stronger than our desired object, and thus even their side lobes can dwarf our objects-of-interest. Fig.~\ref{fig:clutter_source} shows an example of a highly cluttered scene. Our objective is to remove unwanted clutter to focus on the object-of-interest.  While the shape-correlator based detector was designed to avoid ghost peaks, if the object-of-interest is dwarfed by very strong reflections, then these can trigger the correlator detector and result in a faulty depth estimate. Therefore, one must perform a declutter phase prior to applying \name's correlator based peak detection algorithm. Doing so would prevent \name's algorithm from being misled by such strong side lobes.

 
Specifically, in \name\, we look for semantic objects that are usually strong reflectors such as buildings, fences and lamp posts using the camera segmentation output. Using the undesirable reflector's accurate angular locations from the camera, we find if there exist strong peaks outside our angles-of-interest in the radar image. For each such strong peak, we treat it as a point reflector and synthesize a template following Eqn.~\ref{eqn:1}, which captures its contribution to the I/Q signal. A key point to note is that because these are strong reflectors, $\alpha$ of the template is chosen to be equal to the peak value. This template is then subtracted from the received raw I/Q samples. We iterate over different such peaks many times until the magnitude of the peaks in the angles-of-interest are comparable to the expected magnitude of an object reflection. This is analogous to successive interference cancellation in RF communication~\cite{sic}, or the CLEAN algorithm in radio-astronomy \cite{clean}, with the distinction that we only remove the contribution from peaks outside of our angles-of-interest. What this process accomplishes is the removal of side lobes from these large peaks within our angles-of-interest -- thereby enabling robust object peak detection.

\begin{figure*}
\centering
\includegraphics[width=\textwidth]{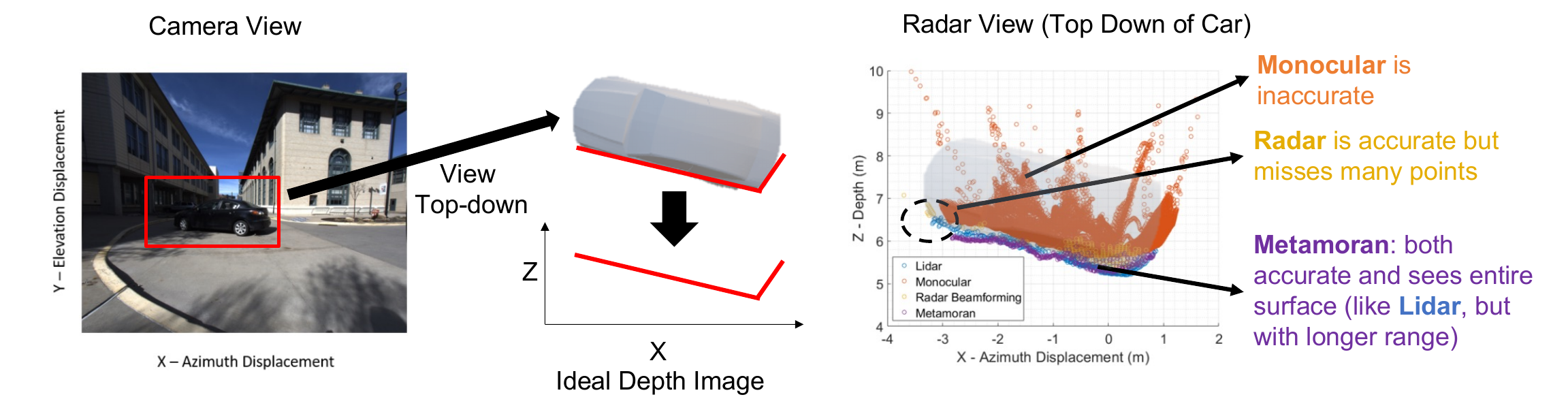}
\vspace*{-0.3in}
\caption{\textbf{\name\ vs. Radar and Monocular Estimation:} A qualitative comparison of the depth images shows standard radar to be very coarse in azimuth resolution, monocular to have significant absolute depth offsets but great azimuth diversity, and \name\ which leverages rich shape information from image pre-processing to generate an accurate, dense depth image.}
\vspace*{-0.1in}
\label{fig:mmwavebeamform}
\end{figure*}

\subsection{Addressing Weak Reflections}  
\label{sec:weakref}
In this section, we explore ways to detect extremely weak reflections from objects-of-interest. Indeed, the precise level to which radar reflections weaken depends on a combination of material properties, poor orientation, small radar cross section, range from the radar and we evaluate this further for a diverse set of objects in Sec.~\ref{sec:microbenchmarks}. 

Our approach relies on the fact that -- because of image segmentation -- we are certain that the object we are looking for exists in a given angular span of interest, and we also know its object type (e.g. car or person). As a result, from historic data, we can determine a received-signal-strength upper bound based on the object type and each distance. Thus, reflectors within the angular span of interest that are significantly higher than expected (and their side lobes) can also be removed as clutter as described in Sec. \ref{sec:clutter} and weak target peaks can then be detected.

While we have so far focused on detecting static objects in high clutter scenarios, we extend \name\ to mobile objects as well by leveraging the Doppler processing that FMCW radars are capable of \cite{tifunda}. With Doppler processing, we show that we can detect and range moving objects even 20dB below the surrounding clutter at ranges of up to 300m (Sec. \ref{sec:dopplerresults}). Depending on how the camera is mounted, it may or may not detect objects up to 300 m. Thus, in this unique case, radar Doppler processing can also inform the camera to pan, tilt and zoom in the right direction and then perform image segmentation to identify the object for surveillance purposes.

%% file: depth_imaging.tex
\section{Depth Imaging} \label{sec:depthimaging}

We note our current description of \name's algorithm provides only one depth value per object template, i.e. one depth per object. In practice, we deal with extended objects and we would require multiple depth values across the object. We could use local peaks from the radar image near the peak depth value obtained from shape-correlation algorithm. But, the point cloud so obtained is very sparse and only becomes sparser with increasing object distances. In an ideal world, we would like an output similar to monocular depth estimation (see Fig.~\ref{fig:monoculardepth} for an example). In monocular depth estimation, pixel color and other image features are used to identify objects at various depth levels resulting in a dense RGB-D image as shown in Fig.~\ref{fig:monoculardepth}. Our key idea is to make use of the dense monocular depth estimation in conjunction with the sparse point cloud obtained from shape-correlator detector output. However two problems persist in realizing this fusion: (1) First, while monocular depth estimation may often correctly return the \textit{relative} depths between different parts of a large object such as a car, it often makes large errors in \textit{absolute} depths, particularly for objects at extended distances, as we note in our experiments in Sec. \ref{sec:depthresults}. (2) Second, monocular depth estimation often struggles with objects that do not have significant variation in color with respect to the background or sharp edges that intuitively simplifies depth estimation~\cite{saxena,farsight}. The rest of this section describes how we address both these challenges to fuse \name's depth images with off the shelf monocular depth estimates that offer superior accuracy to monocular depth estimation.


\begin{algorithm}[!b]
    \SetKwInOut{Input}{Input}
    \SetKwInOut{Output}{Output}
    \caption{Depth Imaging Algorithm}
    \Input{Depth Estimate, $d^{*}$
    \\Contour, $C$ 
    \\Raw I/Q Radar capture, $h$}
    \tcc{Choose local peaks near $d^{*}$ to generate \name's sparse point cloud}
    $MM_{sparse}$ = \textsc{GenerateSparseImage}$(d^{*}, h)$\\
    \tcc{Nullify large absolute errors from monocular estimation}
    $C$ = \textsc{ShiftToDepth}($C$, $d^{*}$)\\
    \tcc{Reject outliers which occur along the edges of the image}
    $C^*$ = \textsc{RejectOutliers}($C$)\\
    $MM_{dense}$ = \textsc{Fuse}($MM_{sparse}, C^*)$
    \hspace*{-0.2in}\Output{$MM_{dense} (x,z)$\hspace*{0.3in}\hfill \tcp{Dense Depth Image}}
    
\end{algorithm}


\sssec{Correcting Absolute Errors: } In Sec. \ref{sec:depth}, we already used monocular depth estimation to generate a contour and found the exact depth at which the contour exists in the radar signal. Because radar's depth estimates are more accurate, we can simply address the first challenge by shifting the contour already generated by the shift estimated by \name's Algorithm-1 from Sec. \ref{sec:depth}.



\begin{figure*}
\centering
\includegraphics[width=.9\textwidth, height=0.15\textheight]{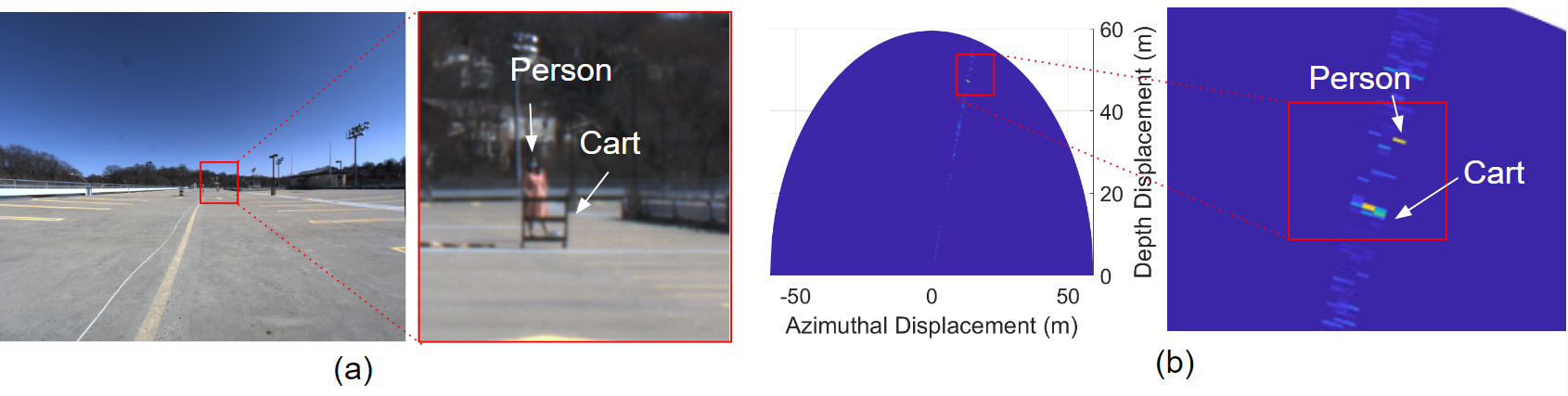}
\vspace*{-0.1in}
\caption{\textbf{\name\ vs. Partial Occlusions:} \name\ can help identify objects-of-interest despite partial occlusions. (a) shows an image of our scene, a person behind a cart, located approximately 45m away. (b) shows radar's capture of the person and the occluding (left) half of the cart. \name\ analyzes image segmentation output to determine which of the two peaks to pick. \name\ then takes the farther reflector as the target.}
\vspace*{-0.1in}
\label{fig:plos}
\end{figure*}

\sssec{Correcting Relative Errors: } After aligning the monocular depth estimates with the sparse point cloud from \name's beamforming, a naive way to fuse this would be consider all points from both modalities. But, as seen in Figure. ~\ref{fig:mmwavebeamform}(b), edges of monocular estimates tend to deviate quite significantly from the primary contour outline of the object. If fused as is, one would experience higher errors as expected from monocular depth estimation. It is therefore important to select points from the aligned monocular depth estimates that only lie along the primary contour outline and reject outliers. We note that the number of points detected per azimuth bin in monocular estimates fall off sharply at the edges where our outliers of interest lie. By using a simple threshold based outlier detection, we identify points which actually lie along the primary contour. We capture the key processing steps in Algorithm-2. Upon fusing selected monocular depth estimate points and sparse point cloud from Sec.~\ref{sec:depth}, we obtain a depth image that outperforms different algorithms using either of the two modalities in terms of azimuth resolution and depth accuracy (see Fig. \ref{fig:mmwavebeamform}).

%% file: partial_occulsions.tex
\section{Partial Occlusions} \label{sec:occlusions}

In this section we describe how \name\ is robust to partial occlusions and vision impediments.

\subsection{Impact of Partial Occlusions}  \label{sec:plos}

We already discussed how to deal with clutter arising from strong reflectors present out of angles-of-interest in Sec. \ref{sec:depthestimation}. Here, we consider another type of clutter originating from objects occluding the object-of-interest. If the object-of-interest was completely occluded by significant blockage, then neither camera nor radar would see it. But in the case of partial obstructions, such as the cart in front of a person pictured in Fig. \ref{fig:plos}, image segmentation will generate a mask for both the obstruction and the object-of-interest. For a known obstruction type, the obstruction can be detected as a target object and then removed as clutter, using techniques explained in Sec. \ref{sec:clutter} and Sec. \ref{sec:weakref}. In the case of an unknown obstruction, we instead fall back to segmentation masks which carry information about what is the foreground and background respectively. If the mask corresponding to the object-of-interest is not distorted then it should be in the foreground, and we choose the closer peak. If the mask is distorted then we choose the peak further away.

\subsection{Vision Impediments} \label{sec:nocamera}

While mmWave radars are known to be fairly resilient to occlusions due to fog~\cite{guan2020through} and lighting conditions, because \name\ uses camera information as prior input, it is critical to think of a system design which works during camera failure as well. In addition to environment and lighting conditions, algorithms such as image segmentation can fail if there is no contrast between the color of the object-of-interest and background. Thus, we build an alternative backup system which kicks in during camera failure.

We do note that most of our algorithms in Sec. \ref{sec:depthestimation} and Sec. \ref{sec:depthimaging} are only possible because of rich camera information that helps us declutter the radar image and provide rich angular information. So in the camera failure mode, we only deal with clutter-free scenes where the object-of-interest is at short ranges (0-20~m). Our task is to detect objects-of-interest in a radar image, classify the object type and choose peaks which provide the sparse depth image. We consider solving this task as drawing a bounding box around a set of peaks corresponding to the object and assigning a label to the box to indicate the object type. In clutter free scenes, any peak is due to an object and thus, detecting is not a huge problem. We then cluster peaks close to the strongest reflector and create the tightest bounding box that covers all of them. We find that identifying objects based on their reflectivities is not ideal. This is because even in clutter-free scenarios the orientation of an object can drastically change the reflected power that radar receives from it (Sec. \ref{sec:microbenchmarks}). Instead, we identify objects based on their physical dimensions. This is already captured in the dimensions of the bounding box. At a fixed range, an extended object like car would have much larger area than a person or stop sign. Because objects like persons and stop signs have similar footprint in a radar image, we are not able to classify between them. We use length and width of the bounding box as features and tune range-specific thresholds that separate car-like and person/stop sign-like objects. We present our results in Sec. \ref{sec:resultsradaronly}.

%% file: results.tex
\begin{figure}
\centering
\includegraphics[width=\columnwidth]{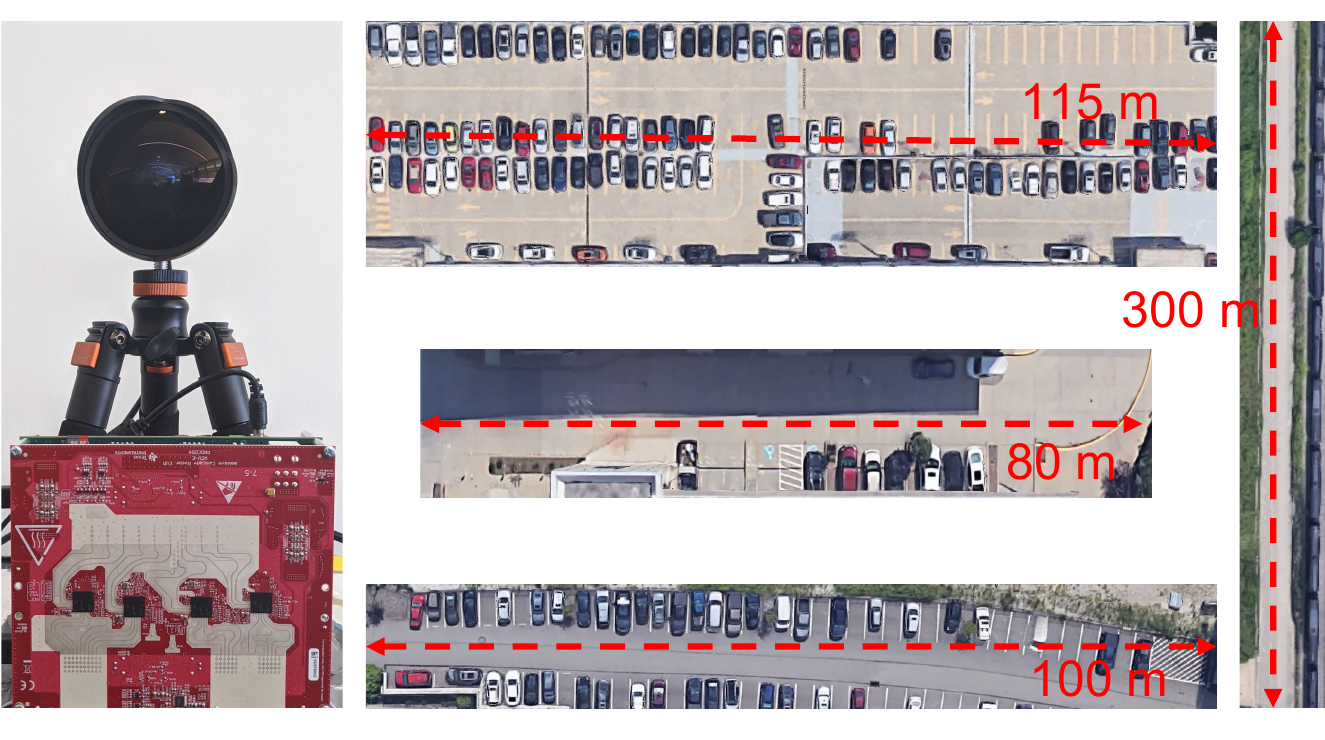}
\vspace*{-0.25in}
\caption{\textbf{\name's Sensing Platform:} We use a FLIR Blackfly S 24.5MP color camera and a TI  MMWCAS-RF-EVM mmWave radar. We deploy our system in outdoor spaces like roads and parking lots with high clutter from buildings, fences, lamp posts, trees, trains, and out-of-interest cars.}
\vspace*{-0.2in}
\label{fig:impl}
\end{figure}

\begin{figure*}[!htb]
    \centering
    \begin{minipage}{0.2\textwidth}
        \centering
        \includegraphics[width=\textwidth]{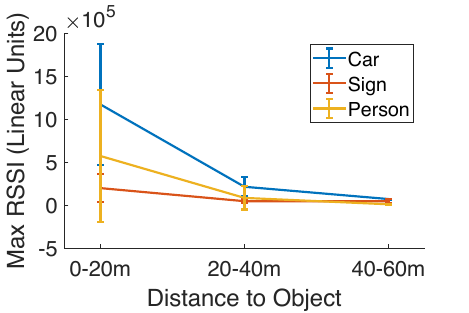}
        \caption{Range Attenuation: Reflectivity of objects in line-of-sight conditions. Variations are due to different orientations.}
        
        \label{fig:micro2}
    \end{minipage}%
    \hspace*{.25cm}
    \begin{minipage}{0.5\textwidth}
        \centering
        \includegraphics[width=\textwidth]{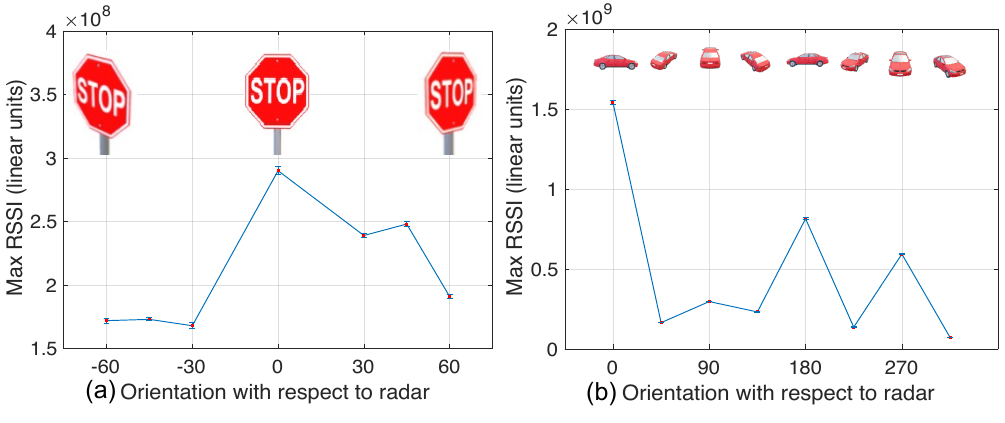}
        \vspace*{-0.25in}
        \caption{Orientation: The magnitude of reflected signal varies with the orientation of our objects-of-interest, stop sign and car.}
        \label{fig:micro1}
    \end{minipage}
    \hspace*{.25cm}
    \begin{minipage}{0.24\textwidth}
        \centering
        \includegraphics[width=\textwidth]{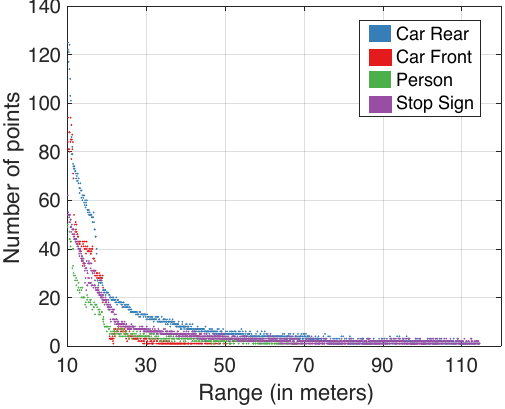}
        \vspace*{-0.2in}
        \caption{Lidar can detect retro-reflective objects such as stop sign up to 115m.}
        \label{fig:lidar_bench}
    \end{minipage}
    \vspace*{-0.1in}
\end{figure*}

\section{Implementation and Evaluation}

\sssec{System Hardware: } \name\ is implemented using a FLIR Blackfly S 24.5MP color camera and a TI  MMWCAS-RF-EVM RADAR (see Fig.~\ref{fig:impl}). We operate the radar at 77-81~GHz with a theoretical range resolution of 3.75-60~cm, depending on max range. The radar also has 86 virtual antennas spaced out along the azimuth axis which provides a theoretical azimuth resolution of $1.4 ^{\circ}$. As explained in Sec. \ref{sec:primer}, this is at least an order of magnitude worse than cameras and lidars. Unlike fusion approaches which rely on processed point clouds~\cite{nobis2019deep}, this radar supports logging raw complex I/Q samples which is critical for our processing.


\sssec{Testbed and Data Collection: } We test this system in a variety of 400 outdoor scenes such as parking lots and roads at distances ranging up to 300~m from objects-of-interest. These environments have rich clutter sources arising due to buildings, street lamps, fences, trees, trains, out-of-interest parked cars and pedestrians. Fig.~\ref{fig:impl} shows four candidate locations in the area surrounding a university campus in a major U.S. city. 


\sssec{Ground Truth: } We collect ground truth data using a Velodyne Puck Lidar(VLP-16), which generates 3D point clouds, with fine angular resolution and 3~cm ranging error. While this lidar is rated for up to 100~m, in practice, on a sunny day, we found the Puck collected data with sufficient point cloud density only until about 20-30~m. Therefore, for ranges beyond 20~m, we surveyed a point closer to the object-of-interest and placed the lidar at that point. 

\sssec{Baselines: } We compare \name\ with two baselines that use the same hardware platforms: (1) \textit{Naive fusion of Camera and Radar: } We use image segmentation to obtain the azimuth spanned by object-of-interest. We perform standard radar beamforming for FMCW radar, and bound the output to the azimuth span and then pick the strongest reflector as the target. (2) \textit{Monocular Depth Estimation: } We use state-of-the-art monocular depth estimation algorithm ~\cite{bhat2020adabins} trained to report depth values up to 80~m. 

\sssec{Objects-of-interest Selection: } We select a car, a person, and a stop sign for use as our objects-of-interest as these are useful for varied applications, including surveillance. Further, these provide a variety of reflectors in size, shape, and reflectivity to evaluate our system. We note that while it is necessary to sense people and cars while they are moving, they are also important to sense when they are stationary -- in the case of a delivery truck, an uber, or a child at a bus stop, for example. Indeed, static objects are much more challenging versus moving objects to detect in radar processing because Doppler-based filtering or background subtraction cannot be used to remove clutter. Hence, the core focus of \name\ is to detect even static objects in high clutter environments. 

\sssec{Calibration: } We note that \name\ requires both internal calibration of the components as well as external calibration between the camera and radar. Internally, our mmWave radar is calibrated using a corner reflector placed at 5~m, as described in the TI's mmWave Studio Cascade User Guide \cite{mmwavestudio}. The camera intrinsics are measured by taking many photos of a checkerboard to remove fisheye distortion (using Matlab's Computer Vision Toolbox~\cite{cvtoolbox}) and for image segmentation and monocular depth estimation. 

Externally, \name\ requires a consistent understanding of object shapes between the mmWave platform system and the camera system. While both of these are co-located in \name, they are at a small relative distance of 15~cm, which could lead to inconsistencies in the images they produce. \name\ accounts for this using a joint calibration of the mmWave radar and camera using a feature-rich metallic surface that is viewed from both the camera and radar platform to capture a Euclidean transform between their frames of reference. The object is chosen to be feature-rich for both platforms, with stark differences in both color and the presence/absence of strong mmWave reflectors (metallic structures). The transform obtained from calibration is applied, prior to fusing measurements from either platform to ensure consistency.






\section{Microbenchmarks} \label{sec:microbenchmarks}

\subsection{Comparing Object Reflectivity}
\label{sec:objectreflectivity}
\sssec{Method: } To determine expected power thresholds for detecting objects-of-interest, we measure the peak value from radar beamforming for our three target reflectors: car, person, and a road sign, across different distances in 81 line of sight settings.

\sssec{Results: } Our results for this are shown in Fig. \ref{fig:micro2}. We observe that power falls off significantly with distance. From about 10~m to 50~m, the reflections attenuate: 16.7$\times$ for a car,  63$\times$ for a person, and 4.4$\times$ for a sign. We note that the sign is a significantly weaker reflector than a person despite being a .762m $\times$ .762m metal sheet outfitted with optical retro-reflectors: past work indicates that this may be due to the majority of incident signal being reflected specularly off planes and thus not received by our radar~\cite{pointillism}.

\subsection{Impact of Object Orientation}
\sssec{Method: } To evaluate the impact of orientation on the reflectivity of our more planar reflectors, we collected data across 7 angles of the front of a stop sign and 8 angles of a car. This data was measured at a fixed 4~m away from the object.

\sssec{Results: } The peak values from radar beamforming at different orientation are shown in Fig. \ref{fig:micro1}. We find that the peaks correspond, as expected, with the largest effective area: the face of the stop sign, and the side of the car. We find the stop sign peak reflectivity degrades 1.68$\times$ at poor orientation, and the car can degrade 21$\times$ depending on orientation. 

\subsection{Lidar benchmark}
\sssec{Method: } To evaluate lidar's maximum detection range and find out the distance at which we have sufficient point cloud density, we collect the lidar data as objects move to far away distances in the same orientation with respect to lidar.

\begin{figure*}[!htb]
\vspace*{-0.1in}
    \centering
    \begin{minipage}{0.3\textwidth}
        \centering
     \vspace*{-0.3in}
        \includegraphics[width=\textwidth]{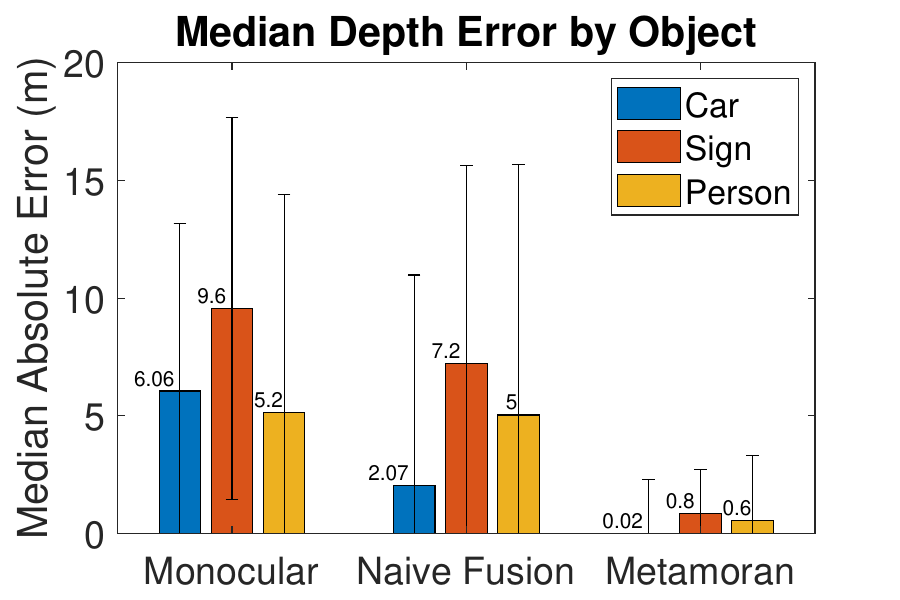}
        \vspace*{-0.25in}
        \caption{We see car with the lowest depth error, followed by person and sign, per their reflectivity.}
        \label{fig:depthalg}
    \end{minipage}%
    \hspace*{.5cm}
    \begin{minipage}{0.3\textwidth}
        \centering
        \includegraphics[width=\textwidth]{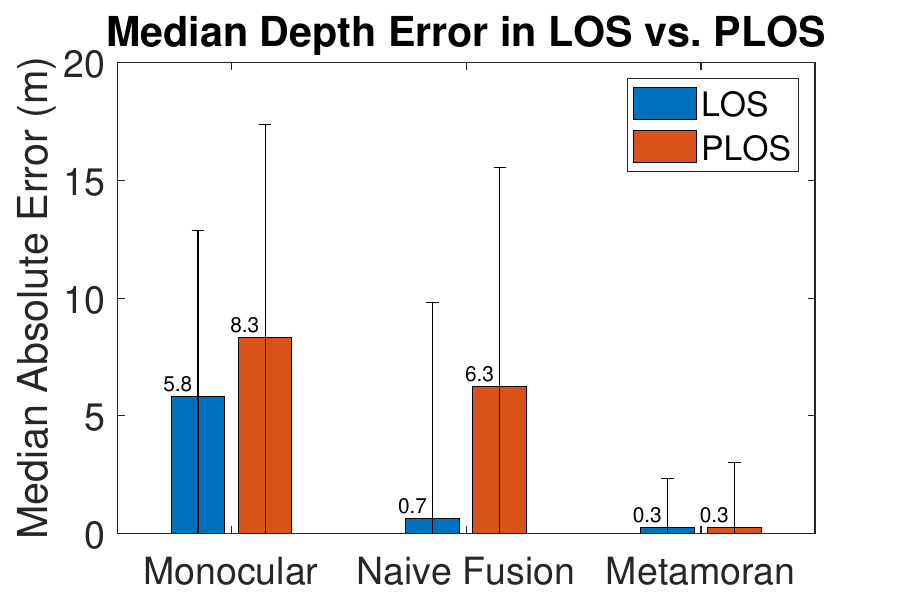}
        \caption{Across all algorithms, we see degraded performance in PLOS compared to LOS, particularly in our naive fusion baseline.\\}
        \label{fig:depthlos}
    \end{minipage}
    \hspace*{.5cm}
    \begin{minipage}{0.3\textwidth}
        \centering
        \includegraphics[width=\textwidth]{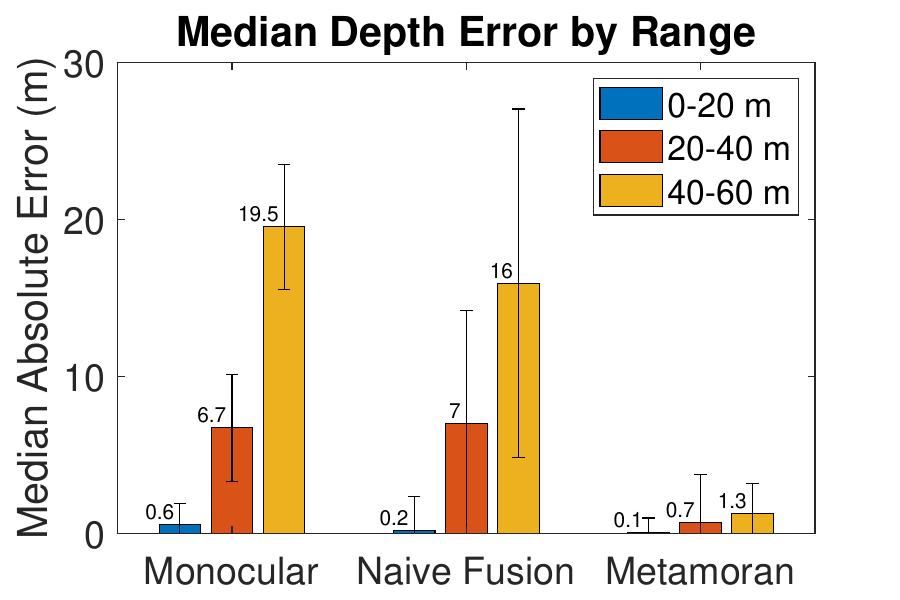}
        \caption{Across all algorithms, we see median depth error rise with increased range, with \name\ showing better accuracy.\\}
        \label{fig:depthrange}
    \end{minipage}
\end{figure*}

\begin{figure*}[!htb]
    \centering
    \begin{minipage}{0.25\textwidth}
        \centering
        \vspace*{-0.3in}
        \includegraphics[width=\textwidth]{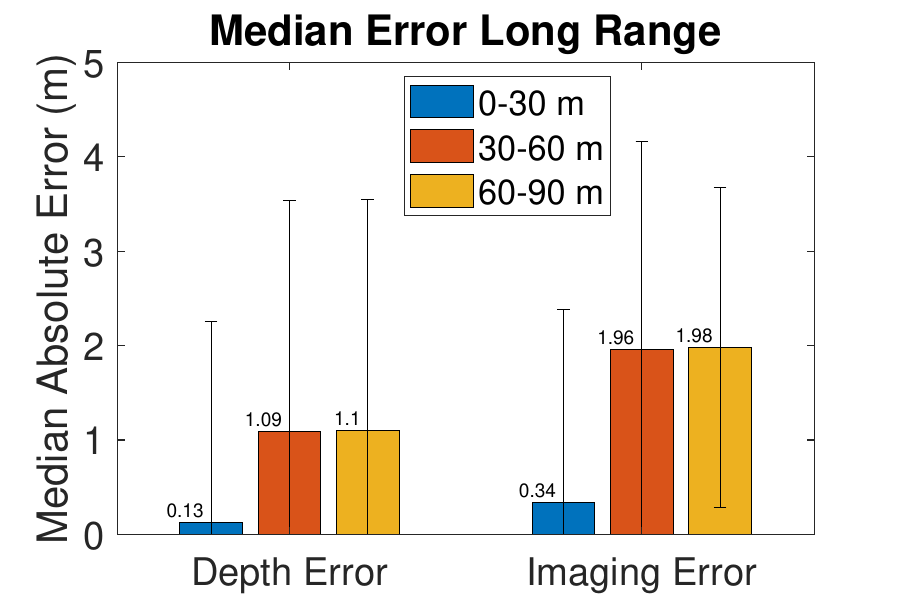}
        \caption{We sense static persons and signs up to 60~m and cars up to 90~m away.}
        \label{fig:rangeext}
    \end{minipage}%
        \hspace*{.5cm}
    \begin{minipage}{0.7\textwidth}
        \centering
        \includegraphics[width=\textwidth]{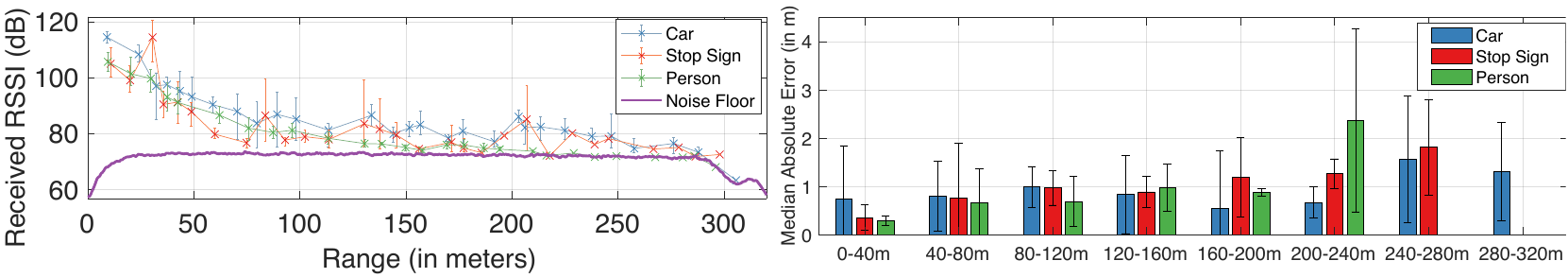}
        \caption{\name\ leverages radar doppler processing to detect and range moving objects even at 300~m where their received RSSI is about 20dB lower than the surrounding static clutter.}
        \label{fig:dop_longrange}
    \end{minipage}
    \vspace*{-0.2in}
\end{figure*}

\sssec{Results: } We noticed that the maximum detection range depends on object reflectivity characteristics. We see in Fig. \ref{fig:lidar_bench} that only one point from the front of the car without a license plate is detected between 30-50~m. Depending on the color of the paint and orientation of the car, we observed that the front would stop being detected even between 25-30~m. This could largely be because of mirror-like reflectivity causing reflections to never return to lidar. However, objects with retroreflective surfaces such as rear of the car with license plate and stop sign are detectable up to 114.3~m and 114.6~m respectively. A person being a diffuse scatterer is detectable up to 64~m. Surveillance applications cannot afford to make any assumptions on the mirror like/retroreflectivity of objects. Although, specular properties have been investigated in radar \cite{pointillism}, we show that without making any reflectivity assumption \name\ can detect all orientations of car, stop sign and person effectively. However, point cloud density drops drastically as objects move away. We pick 20~m as the range with sufficient point cloud density. For collecting ground truth beyond 20~m, we survey and move the lidar to a point closer to the object. 

\section{Results} \label{sec:results}

\subsection{Depth Estimation}
\label{sec:depthresults}


\sssec{Method: } For range results, we collected 146 data samples in varying lighting conditions at 2 clutter rich sites. We collected both line-of-sight (LOS) captures of targets and captures of partial line-of-sight (PLOS) occluded by carts. Targets were positioned from 3-58~m and were placed in various orientations with respect to radar. Data was collected in 2 range/resolution buckets: 4.2~cm at 0-20~m, 11.6~cm at 20-60~m. The primary bottleneck of range resolution for this system is the TDA2SX SoC capture card  on the MMWCAS board -- it can handle at most a data width of 4096, corresponding to 512 complex samples per receiver. Thus at longer ranges, we can't utilize the full potential of mmWave range resolution. 

Depth error is measured at the same point in each of these approaches (Peak value obtained with naive fusion of radar beamforming and camera, depth prediction from monocular depth estimation, and Metamoran estimate) to the depth provided by the lidar. We compare median error in depth across objects-of-interest for \name\ and the two baseline systems. We include error bars corresponding to +/- the standard deviation of our collected data. It should be noted that we present median over mean due to the long tail often found in RF localization and sensing that affects both \name\ and baselines: slight variances in noise and power can result in disproportionately large errors if the second-largest peak overtakes the first. For systems with a low median error, this effect can be ameliorated by taking multiple snapshots and removing outliers. Because of this long tail, we do notice that our standard deviations tend to be large.

Below, we represent three sets of results: (1) three different objects; (2) partial occlusions preventing a complete direct view of the object; (3) three different range buckets. Across all experiments, we find that \name\ significantly outperforms the baselines.

\sssec{Object Results: } Fig.~\ref{fig:depthalg} shows the median error in depth across objects-of-interest for \name\ and the two baseline systems. We see lowest error for the car across the board due to a combination of factors: the car is our strongest reflector, offers multiple points on its surface to reflect radar signals due to its size and thereby a high radar cross section. We see performance further degrade with the progressively weak reflectors as measured in Sec. \ref{sec:objectreflectivity}: the person is the next most accurate, followed by the sign.

\sssec{Occlusion Results: } Fig.~\ref{fig:depthlos} shows the median error in depth in line-of-sight (LOS) and partial-line-of-sight (PLOS) for \name\ and the two baseline systems. We see significant degradation in our naive fusion baseline for PLOS, which frequently takes the occluding object as the strongest reflector, unlike \name, which can detect and account for occlusions using techniques in Sec. \ref{sec:plos}. 

\sssec{Range Results: } Fig.~\ref{fig:depthrange} shows the median error in depth across range for \name\ and the baselines. As expected, accuracy across all approaches, objects, and occlusion settings deteriorates with range due to weaker received signals. 

\begin{figure*}[!htb]
    \centering
    \begin{minipage}{0.25\textwidth}
        \centering
        \includegraphics[width=\textwidth]{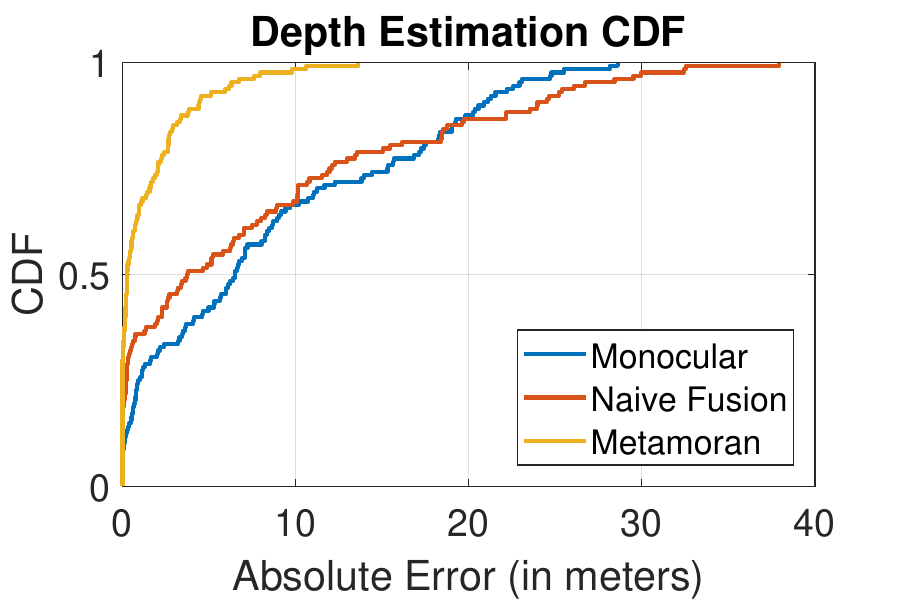}
        
        \caption{CDF of depth errors shows our depth estimation is accurate even in high clutter. \\ 
        }
        \label{fig:depthcdf}
    \end{minipage}%
    \hspace*{.5cm}
    \begin{minipage}{0.25\textwidth}
        \centering
        \includegraphics[width=\textwidth]{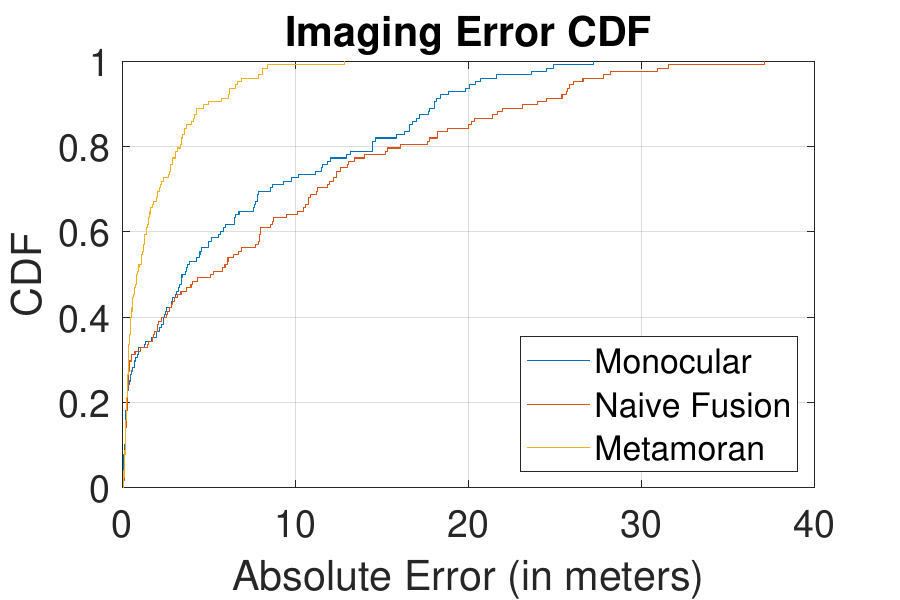}
        \caption{CDF of depth imaging errors shows our imaging is accurate even in high clutter \\}
        \label{fig:shapecdf}
    \end{minipage}%
    \hspace*{.5cm}
    \begin{minipage}{0.4\textwidth}
        \centering
        \includegraphics[width=\textwidth] {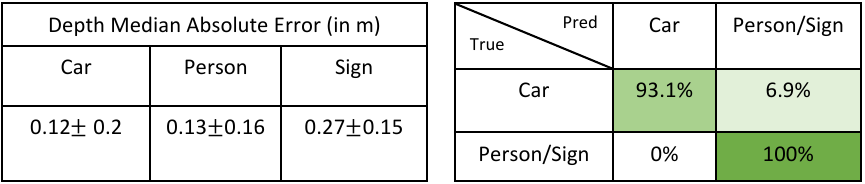}
        \caption{In scenes where camera processing fails, but is clutter-free with object-of-interest in short ranges, we perform depth estimation and classification using radar alone.}
        \label{fig:table}
    \end{minipage}%
    \vspace*{-0.3in}
\end{figure*}

\sssec{CDF Results: } Fig.~\ref{fig:depthcdf} shows CDF of the median error in depth for \name\ and the baselines. \name\ has a median error of \textbf{0.28~m} across all collected data, compared to 6.5~m for monocular depth estimation and 3.75~m for naive radar and camera fusion. These correspond to mean values of 1.42~m, 8.48~m, and 7.89~m respectively due to long tail effects. \name\ clearly outperforms the baselines to accurately detect a variety of objects in different orientations and in high clutter environments.

\begin{figure*}[!htb]
    \vspace*{-0.3in}
    \centering
    \begin{minipage}{0.3\textwidth}
        \centering
        \includegraphics[width=\textwidth]{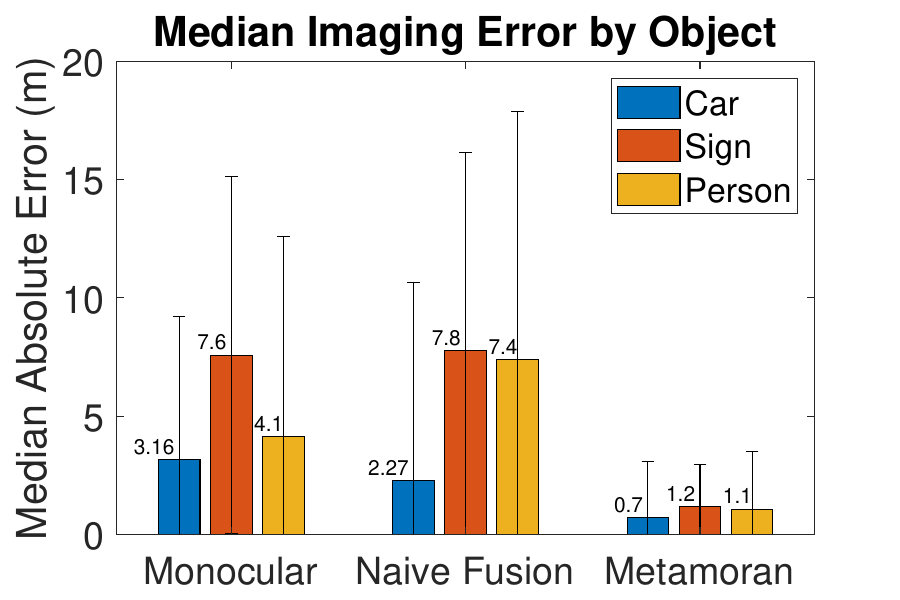}
        \caption{Imaging Errors increase with decreasing object reflectivity across algorithms.}
        \label{fig:shapealg}
    \end{minipage}%
    \hspace*{.5cm}
    \begin{minipage}{0.3\textwidth}
        \centering
        \includegraphics[width=\textwidth]{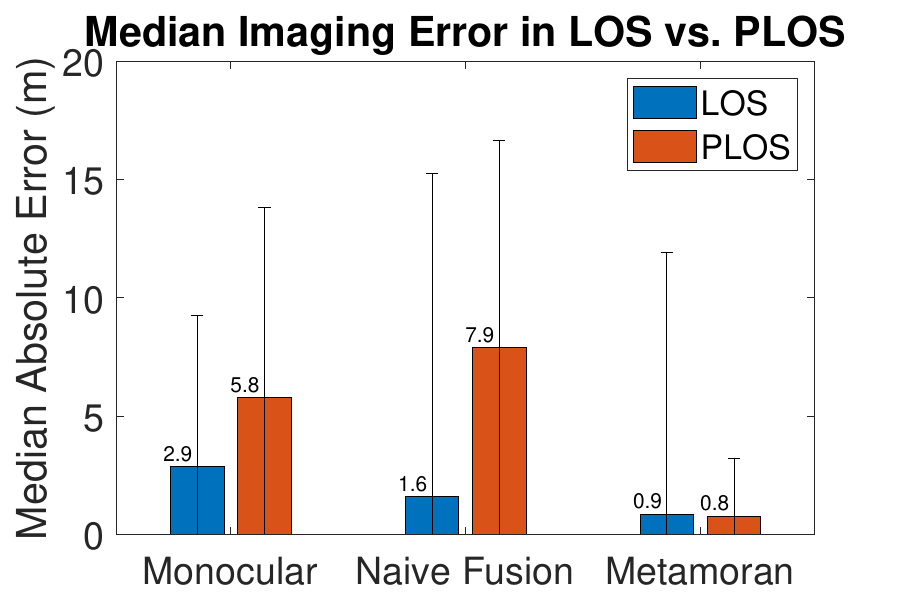}
        \caption{Imaging Errors are degraded in partial line of sight scenarios across all algorithms. }
        \label{fig:shapelos}
    \end{minipage}
    \hspace*{.5cm}
    \begin{minipage}{0.3\textwidth}
        \centering
        \includegraphics[width=\textwidth]{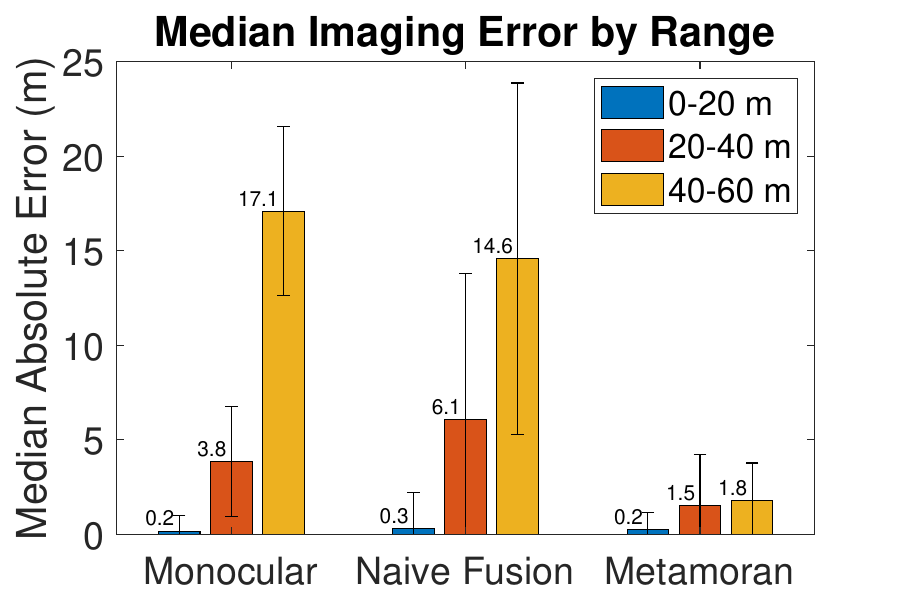}
        \caption{Imaging Errors increase with increasing range.\\ }
        \label{fig:shaperange}
    \end{minipage}%
\end{figure*}

\subsection{Extremely Long Ranges} \label{sec:dopplerresults}
\sssec{Method: } To evaluate the maximum system range of \name\ we perform the following experiments beyond 60~m. At these extended ranges, our baselines either no longer detect objects or encounter huge errors. Sign and person are no longer detectable even with the assistance of \name. We collect data at 2 sites for static cars beyond 60~m. We also leverage the Doppler processing that FMCW radars are capable of and to evaluate the true radar detection ability we collect data for moving car, person and stop sign up to 320~m. For these ranges, we collect the data at 30~cm range resolution up to 120~m and 60~cm resolution up to 320~m.

\sssec{Results: } We find that static cars are no longer detectable beyond 90~m. We show the results up to 90~m for depth estimation of \name\ compared to the lidar ground truth in Fig. \ref{fig:rangeext}. We see slight degradation with the increased distance. Next, for moving objects, we see in Fig. \ref{fig:dop_longrange} that the received signal strength drops consistently until 305~m, when it hits the noise floor. The signal strength variations are particularly large for stop sign because they are sensitive to orientation with respect to radar. The person is detected up to 229~m, stop sign up to 298~m and car up to 305~m. Although the reflection from these objects at long ranges are extremely small versus background clutter, just because they are moving, Doppler processing can still detect the objects. We also see that in Fig. \ref{fig:dop_longrange} the depth errors increase with distance as expected. Because of the radar resolution of 60~cm, at these long ranges, even for cars the errors can reach 1.5~m. Given enough signal integration, these errors should decrease and reach the resolution limit. As long as the radar detects these objects, \name's depth imaging algorithms are still applicable albeit with the help of a pan, tilt, zoom camera to get a high resolution image of the object for fusion.


        


\subsection{Depth Imaging}\label{sec:resultintra}

\sssec{Method: } To compute high resolution depth images, we implement the method in Sec. \ref{sec:depthimaging}. In contrast to Sec. \ref{sec:depthresults} which only computed depth errors, here we want to characterize system performance for a point cloud obtained from the baselines monocular depth estimation and naive fusion of camera and radar, and our system against lidar point clouds. Data was collected similar to Sec. \ref{sec:depthresults}. 

To compare two point clouds $A$ and $B$, we use a modified version of Hausdorff distance \cite{hausdorff} as follows:
\begin{equation*}
    \operatorname{min}
    \Bigg\{\underset{a \in A}{\operatorname{median}} \big\{\underset{b \in B}{\operatorname{min}} \{d(a,b)\} \big\}, \underset{b \in B}{\operatorname{median}} \big\{\underset{a \in A}{\operatorname{min}} \{d(b,a)\} \big\} \Bigg\} 
\end{equation*}
where $d(a,b)$ is the distance between points $a$ and $b$. Hausdorff distance is popularly used in obtaining similarity scores between point clouds. Intuitively, this metric measures the median distance between any two points in the point cloud. The lower the distance, the more similar the point clouds are. We report this distance as imaging error in meters.


\sssec{Results: } Trends in imaging results largely follow those in depth imaging, as problems with detection propagate through the system. We note that imaging error is larger than the depth error across the board due to additional pairwise distances being calculated. Fig. ~\ref{fig:shapealg} shows the imaging errors against different object types for the 3 different algorithms, Fig. ~\ref{fig:shapelos} shows the median error in imaging in line-of-sight and partial-line-of-sight for \name\ and the two baseline systems, and Fig. ~\ref{fig:shaperange} shows the median error in depth across range for \name\ and the two baseline systems. \name\ outperforms both baselines across all categories. Fig.~\ref{fig:shapecdf} shows CDF of the median error in depth for \name\ and the two baseline systems. \name\ has a median error of \textbf{0.8m} across all collected data, compared to 3.4~m for monocular depth estimation and 5.04~m for naive radar and camera fusion. These correspond to mean values of 1.82~m, 6.59~m, and 8.27~m respectively due to long tail effects. We note that in these baselines, monocular depth estimation outperforms naive fusion unlike in Sec. \ref{sec:depthresults}. This is because monocular depth estimation benefits from our metric due to its large azimuth span of many points that are thus more likely to be close to a point in the lidar, versus the fewer, and sparse points given by naive fusion.

\subsection{Vision Impediments} \label{sec:resultsradaronly}
\sssec{Method: } We introduced our all-radar system in Sec. \ref{sec:nocamera} to deal with events of complete camera failure. This alternative system is only capable of operating in no clutter, short ranges of 0-20~m. We once again collect radar data for car, person and stop sign, at high range resolution of 4.2~cm using the full bandwidth available.

\sssec{Results: } Fig. \ref{fig:table} depicts the depth estimation errors and classification accuracy of radar-only \name. We see that the depth estimation is robust because without clutter, any peak corresponds to the object and radar beamforming is good enough to detect these. We also show high classification accuracy between the car and person/sign. Note that because our technique uses object size as feature, we cannot distinguish between person and sign. A few orientations of a car can create a small signature on radar image and our technique misclassifies these as person/sign.

%% file: limitations.tex
\section{Limitations}\label{sec:limitations}

An important feature of our system is its reliance on camera which provides highly useful input to declutter the scene and detect far away objects. However, camera performance is sensitive to occlusions due to weather such as fog, lighting conditions and lack of contrast between color of the object-of-interest and background. In the event of camera failure, although our alternative system kicks in, we are not capable of disambiguating clutter from object at long ranges in high clutter environments.




\section{Conclusion}

This paper develops \name, a hybrid mmWave and camera based system that achieves high resolution depth images for objects at long ranges and in high clutter environments. \name's secret sauce is in leveraging processed camera information to declutter the scene, eliminate false peaks and identify the right peaks. \name\ also uses the detected peak and processed camera information to create a high resolution depth image of objects-of-interest. \name\ was evaluated extensively up to 300~m. The resulting dataset is extremely valuable to the community as it offers ground truthed lidar, camera and raw I/Q radar data. We believe there is a strong role for radar and camera to play for deploying surveillance applications and that there is rich scope for future work by extending and evaluating on broader classes of objects and ensuring resilience to severe occlusions.